\pdfoutput=1

\documentclass[11pt]{article}

\usepackage{ACL2023}

\usepackage{times}
\usepackage{latexsym}
\usepackage{subfig}
\usepackage{times}
\usepackage{latexsym}
\usepackage{amsmath}
\usepackage{dashrule}
\usepackage{graphicx}
\usepackage[ruled,linesnumbered]{algorithm2e}
\definecolor{mydarkblue}{rgb}{0,0.08,0.45}
\usepackage{stmaryrd}
\usepackage{float}
\usepackage{footnote}
\usepackage{enumitem}
\usepackage{bm}
\usepackage{arydshln}
\usepackage{booktabs}
\usepackage{multirow}
\usepackage{color}
\usepackage{comment}
\usepackage[font=small]{caption}
\usepackage{fancyhdr}
\usepackage{xspace}
\usepackage{arydshln}
\usepackage{graphicx} 
\usepackage{booktabs}
\usepackage{multirow,multicol}
\usepackage{amsmath,amssymb}
\usepackage{enumitem}
\usepackage{fancyvrb}
\usepackage{wrapfig}
\usepackage{nameref}
\usepackage{mdframed}
\usepackage{comment}
\usepackage{titlesec}
\usepackage{mathtools}

\usepackage[T1]{fontenc}

\usepackage[utf8]{inputenc}

\usepackage{microtype}

\usepackage{inconsolata}

\usepackage{tikz}
\usetikzlibrary{positioning}
\usepackage[most]{tcolorbox}

\usepackage{etoolbox}
\usepackage{diagbox}
\AtBeginEnvironment{tcolorbox}{\small}
\usepackage{pdfpages}

\usepackage{color}
\usepackage{colortbl}
\definecolor{intnull}{RGB}{225, 245, 164}
\definecolor{intyellow}{RGB}{250, 244, 135}
\definecolor{intgray}{RGB}{233, 239, 240}
\definecolor{intred}{RGB}{245, 224, 201}
\definecolor{patterngreen}{RGB}{102, 212, 149}
\definecolor{semanticblue}{RGB}{152, 227, 226}

\definecolor{myblue}{rgb}{0.82, 0.94, 0.75}
\definecolor{mygreen}{rgb}{0.64, 0.76, 0.68}
\definecolor{myyellow}{rgb}{0.88, 0.54, 0.35}
\definecolor{mygreen}{rgb}{0.68, 0.9, 0.8}
\definecolor{mypink}{rgb}{0.2, 0.87, 0.2}
\def\arrvline{\hfil\kern\arraycolsep\vline\kern-\arraycolsep\hfilneg}

\def\mystrut(#1,#2){\vrule height #1pt depth #2pt width 0pt}   

\definecolor{purple}{rgb}{0.5,0,1}
\definecolor{dcyan}{rgb}{0.2,0.6,0.5}
\definecolor{light-gray}{gray}{0.95} 

\definecolor{darkgreen}{RGB}{0,140,0}
\definecolor{darkred}{RGB}{200,0,0}
\definecolor{lightgreen}{RGB}{197, 237, 208}
\definecolor{lightred}{RGB}{255,205,212}
\definecolor{lightyellow}{RGB}{255,240,160}
\definecolor{lightblue}{RGB}{195,221,255}
\definecolor{lightpurple}{RGB}{232,209,255}
\definecolor{lightgray}{RGB}{205,205,205}
\definecolor{indigo}{RGB}{13,165,240}

\usepackage{inconsolata}

\newcommand{\name}{\textsc{AnaloBench}}

\newcommand{\ie}{\emph{i.e.}}

\newcommand{\LLaMATwo}{LLaMA2}
\newcommand{\GPTFour}{GPT-4}
\newcommand{\GPTThreeFive}{GPT-3.5}
\newcommand{\UnifiedQA}{UnifiedQA}
\newcommand{\XwinLM}{XwinLM}
\newcommand{\WizardLM}{WizardLM}

\newcommand{\TuluTwo}{Tulu2}
\newcommand{\Zephyr}{Zephyr}
\newcommand{\ClaudeVTWo}{Claude-v2}

\newcommand{\TOne}{$T_1$}
\newcommand{\TTwo}{$T_2$}

\def\sectionautorefname~#1\null{\S#1\null}
\def\subsectionautorefname~#1\null{\S#1\null}
\def\subsubsectionautorefname~#1\null{\S#1\null}

\title{
\vspace*{-0.5in}
{{\small \hfill EMNLP'24}\\
\vspace*{.25in}} 
{\name:} Benchmarking the Identification of\\Abstract and Long-context Analogies}

\usepackage{tikz}
\newcommand*\circled[1]{\tikz[baseline=(char.base)]{
            \node[shape=circle,draw,inner sep=0.6pt] (char) {#1};}}
\usepackage[textsize=tiny]{todonotes}

\newcommand{\storyanalogy}{\textsc{StoryAnalogy}}
\newcommand{\gentner}{\textsc{Gentner}}

\author{
Xiao Ye$^{\heartsuit}$ \quad 
Andrew Wang$^{\heartsuit}$ \\
{\bf 
Jacob Choi \quad 
Yining Lu \quad
Shreya Sharma \quad 
Lingfeng Shen \quad 
Vijay Tiyyala}\\
{\bf
Nicholas Andrews \; 
Daniel Khashabi
}\\
Johns Hopkins University\\
\texttt{\{xye23, awang116, danielk\}@jhu.edu}
}

\begin{document}
\maketitle
\def\thefootnote{$\heartsuit$}\footnotetext{
Co-first authors
}
\def\thefootnote{\arabic{footnote}}
\begin{abstract}
Humans regularly engage in analogical thinking, relating personal experiences to current situations ($X$ is analogous to $Y$ because of $Z$).
Analogical thinking allows humans to solve problems in creative ways, grasp difficult concepts, and articulate ideas more effectively. 
Can language models (LMs) do the same? 
To answer this question, we propose \name,
a benchmark to 
determine analogical reasoning ability in LMs.
Our benchmarking approach focuses on aspects of this ability that are common among humans:
(i) recalling related experiences from a large amount of information, and (ii) applying analogical reasoning to complex and lengthy scenarios.
We collect a set of 340 high quality, human written analogies for use in our benchmark, which constitutes the largest such collection to date.
We then test a broad collection of models consisting of 12 open source and 3 proprietary in various sizes and architectures. 
As in prior results, scaling up LMs results in some performance boosts. Surprisingly, scale offers minimal gains when, 
(i) analogies involve lengthy scenarios, 
or 
(ii) recalling relevant scenarios from a large pool of information, a process analogous to finding a needle in a haystack. 
We hope these observations encourage further research in this field.\footnote{Code and data  is available online: \url{https://github.com/JHU-CLSP/AnaloBench}}
\end{abstract}

\section{Introduction}
\label{sec:intro}

Analogy is the ability to think about relational patterns~\citep{holyoak2001place} and forms an integral aspect of human communication~\cite{hofstadter2001analogy,gentner2017analogy}. 
This cognitive ability helps humans understand new or difficult concepts by relating them to more familiar experiences \cite{holyoak1996mental}.
Analogical thinking plays a critical role in 
some of the major breakthroughs in human history, such as the discovery of gravity or even Einstein's theory of relativity~\citep{hesse1965models,stepan1986race,hofstadter2013surfaces}.
It was this very analogy-driven progress that Newton aptly described as \emph{``standing upon the shoulders of giants,''} itself an analogy. 
If modern language models (LMs)~\cite{openai2023gpt4,touvron2023llama2} can leverage analogical thinking, then we can expect wide-ranging implications for future tasks. 

\begin{figure}[t]
    \centering
    \includegraphics[scale=0.99,trim=1.5cm 5.3cm 16.5cm 0.4cm,clip=true]{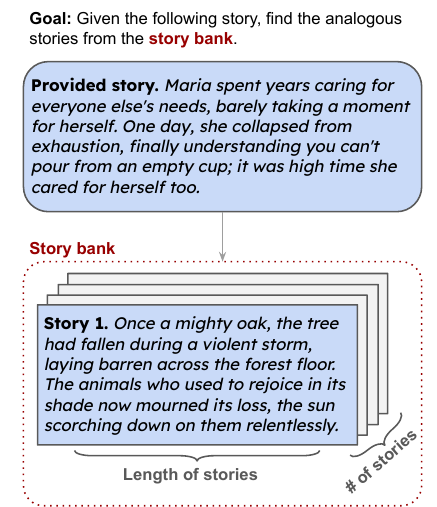}
    \caption{
        The problem setup: given a story, the goal is to identify an analogous story from a story bank. 
        We study the difficulty of this goal for LMs by varying the following parameters: 
        (i) length of stories, (ii) number of stories in the story bank. In the example, both ``Maria'' and ``the oak'' lose the ability to provide for others. While the strength of analogies can vary, we design our benchmark to account for this variation.
    }
    \label{fig:teaser}
\end{figure}

\begin{figure*}[ht]
    \centering
    \includegraphics[scale=.65, trim=.65cm 4.25cm 0.2cm .8cm,clip=true]{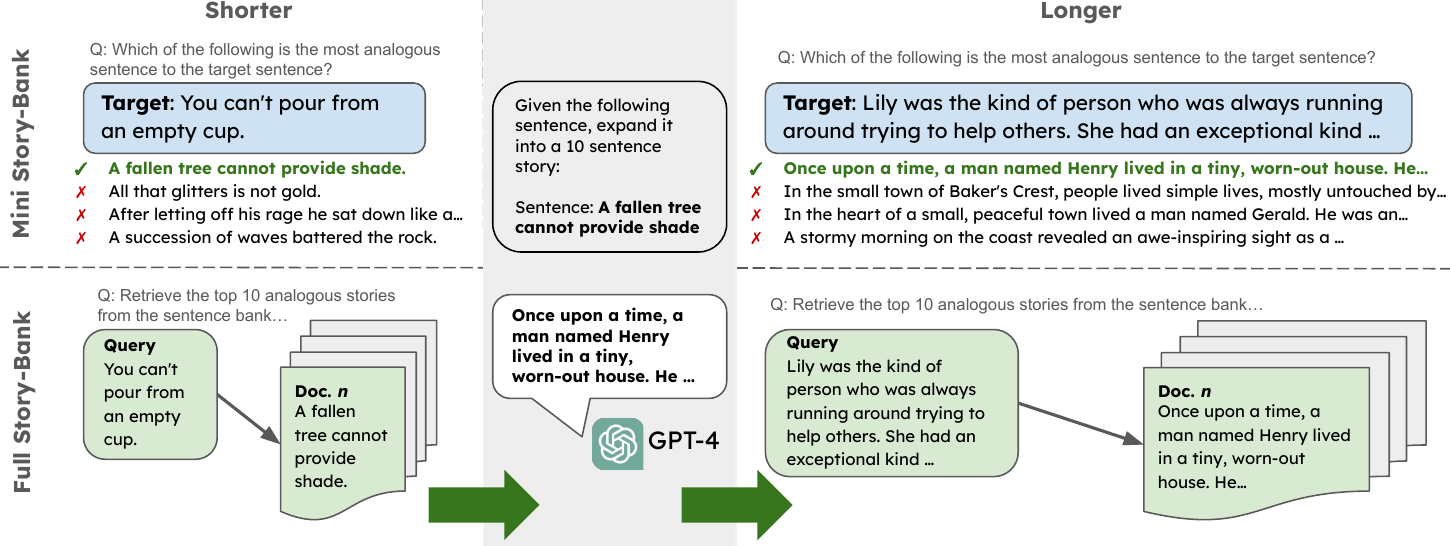}
    \caption{Overview of \name, for both the story expansion \circled{3} and the task creation \S\ref{sec:task-descriptions}. Our abstract analogy identification benchmark features two tasks: (\TOne) Identifying analogies from a mini story bank and (\TTwo) Identifying analogies from a large story bank. Each task is repeated at varying story lengths ($\sim$ 1, 10, and 30 sentences), with LLMs extending each story to target length. We find that while analogical reasoning shows signs of emergence, reasoning over longer and more complex analogies remains a challenge for state of the art LMs. }
    \label{fig:intro-figure}
\end{figure*}

We assess the ability of LMs to handle components of analogy making.
Two important features characterize how humans form analogies in creative pursuits.
(1) Humans are able to pinpoint analogies between prolonged experiences (e.g. ``obtaining a PhD is like running a marathon'').
(2) Humans can recollect relevant analogs from a large collection of past experiences to form analogies \cite{keane1987retrieving,wharton1994below}.
To what extent are LMs capable of such abilities?

To answer the above questions,  we introduce \name, a benchmark for analogical reasoning over natural language stories that convey abstract concepts with varying level of difficulty. 
While the dominant treatment of analogies has been limited to word-level lexical analogies\footnote{e.g. ``rock'' is to ``solid'' as ``water'' is to ``liquid''}~
\cite{mikolov2010recurrent}, 
we instead focus on analogies defined on natural language documents, such as the one shown in Fig.~\ref{fig:teaser}.
In the example, the central figure of each stories (Maria / the ``mighty oak'') loses the ability to provide for the needs of others (``collapsed from exhaustion'' / ``the tree had fallen''). The use of stories as components of analogies provides a natural way to introduce abstract relational patterns. In total, we collect 340 pairs of high-quality analogous stories from human annotators 
after multiple rounds of review and editing.

As Fig.~\ref{fig:teaser} shows, we are 
interested in quantifying the extent to which LMs are capable of identifying analogous stories from a given pool of candidate stories, similar to humans' ability to recollect past experiences and relate them to new situations.   
We characterize this goal with two tasks (\S\ref{sec:task-descriptions}). 
First, we consider a setup where the pool is limited to a few stories.
Among these few candidates, the model is expected to select exactly one story as the closest analogy to a given story ($\emph{\textbf{T}}_1$). 
Good performance requires demonstrated ability in identifying complex analogies, assuming a small pool of candidates. 
In our second task, we maintain a large ($\approx$ 200) pool of candidate stories ($\emph{\textbf{T}}_2$) --- in performing well on this task, a model will have demonstrated ability in identifying analogies from long-context memory. 
Additionally, we explore how well performance scales with length. We are inspired by the remarkable ability of humans to abstract over long and elaborate stories, and leverage such abstractions to identify analogies.
By evaluating our proposed tasks on longer stories, we measure the extent LMs can abstract over complexities of longer stories.
In practice, we repeat each experiment with the same stories told using $\approx$ 1 sentence, 10 sentences, and 30 sentences. 
We benchmark existing open-source and private language models to measure their ability to identify abstract and long-context analogies (\S\ref{sec:experiments}). 
We find that, while scaling LMs leads to better performance in 1-sentence stories, the gains afforded by scale is minimal for longer stories. 
Furthermore, the gap between humans and GPT4 is 6.9\% 
on 1-sentence stories, but increases to 28.8\%
on 30-sentence stories, 
demonstrating that long and complex analogies pose a challenge for LMs.

In summary, we introduce \name~(\autoref{fig:intro-figure}), a novel benchmark with two analogical reasoning tasks, 
and provide a thorough analysis of analogical reasoning ability in a wide range of state of the art language models.
\section{Related Work}

\paragraph{Analogical reasoning datasets.}
Various efforts have attempted to build analogical reasoning benchmarks. 
Within the AI literature, the majority of these works focus on lexical analogies  (\ie, \emph{``man'' to ``woman''} $\approx$ \emph{``boy'' to ``girl''})~\cite{sternberg1980developmental,turney2008uniform,green2012neural,jurgens2012semeval,mikolov2013distributed,mikolov2013linguistic,gladkova2016analogy,lu2019emergence,ushio2021bert}. 
Most of these datasets are created manually, although there are also lexical analogy resources that are created semi-automatically. For example, \citet{yuan2023analogykb} presents a dataset with over a million lexical analogies derived from a knowledge base of subject-object-verb triplets. However, lexical analogies fail to properly test reasoning ability in LMs \citep{yuan2023beneath}.
More recently, research has turned towards proverbs and metaphors for richer analogy benchmarks \citep{ghosh2022epic, wijesiriwardene2023analogical}. Yet proverbs and metaphors no longer challenge modern LMs, with datasets such as ePiC \citep{ghosh2022epic} excluded from Big Bench Hard for this reason \citep{suzgun-etal-2023-challenging}.
Our work ventures beyond lexical analogies and focuses on \textit{challenging analogies} that involve paragraphs of raw-form text, without any assumptions on their structure.
 
Another group of datasets are from cognitive science, some of which involve long sentences. 
These datasets were originally intended to be used for the study of analogical reasoning in humans~\cite{gick1980analogical,keane1987retrieving,gentner1993roles,weinberger2016conscious}. The majority of these datasets are too small to provide reliable benchmarking for models.  
Among these \gentner~\cite{gentner1986systematicity} contains 54 instances and was created to examine the development of systematicity (i.e., sensitivity to parallels based on more complex relations). 
Recently, \cite{webb2023emergent} observes strong performance of LLMs on these datasets, which motivates introducing a more challenging analogical reasoning benchmark.

Concurrent works include \storyanalogy~ \citep{jiayang2023storyanalogy}, a benchmark of 24K sentence pairs, which were generated semi-automatically using GPT-3 and then relabeled by human annotators, and ParallelPARC \citep{sultan2024parallelparc}, a set of 4288 machine generated analogies with a subset of 310 verified by humans.
Compared to these works, our benchmark is much smaller as we prioritize data quality over size (\autoref{appendix:seed-analogies}). Our seed data is all written by humans, at the cost of size, mainly because we aimed at effective evaluation. 
Other works derive benchmarks from established data sources. ARN \citep{sourati2024arnanalogicalreasoningnarratives} constructs analogies between stories in ePiC, using shared proverbs as a proxy for shared relational structure. Unlike ARN, we contribute an entirely new set of 340 seed stories for future work, and propose a different method for coming up with narratives. Furthermore, we evaluate the effect of story length on model performance. 

It is worth noting that there is also a literature on \emph{visual} analogies~\cite{sadeghi2015visalogy,bitton2023vasr,reed2015deep,zhang2019raven} that is different from the scope of this work. 
Interested readers can refer to \citet{ichien2020verbal} who provide a thorough review of the prior datasets both in computer science and cognitive science literature.

\paragraph{Analogical reasoning in LMs.}
Since the rise of pre-trained LMs, we have witnessed remarkable gains in the abilities of these models in tackling analogical 
reasoning~\cite{ichien2023large,webb2023emergent}. Even without using SOTA LMs, \citet{sultan2023life} demonstrated that analogies could be mined and retrieved successfully from a set of situations.
\citet{bhavya2022analogy} studied the ability of GPT3 in generating analogous statements with prompting by  literal mentions of ``analogy'' in prompts. 
Through crowdsourcing experiments, they observe that the then largest models (e.g., \texttt{davinci})
were able to generate analogies that matched the quality of human-generated analogies. 
Another remarkable milestone is reported by~\citet{webb2023emergent} who evaluate GPT3 on various analogical reasoning tasks (Raven’s standard progressive matrices, letter string analogies, etc.) and 
report that ``GPT-3 displayed a surprisingly strong capacity for abstract pattern induction, matching or even surpassing human capabilities in most settings.''
While our results align with these findings, 
our benchmark reveals major limitations of LMs that was not easily observable in the prior work (e.g., the weakness of LMs in solving analogies that involve longer inputs).

\section{\name: A Benchmark for Abstract and Long-Context Analogies}
\label{sec:benchmark}

We discuss design considerations and challenges of benchmarking analogies
(\S\ref{subsec:desiderata}), the construction of \name{} (\S\ref{subsec:creation}), and tasks devised based on this dataset (\S\ref{sec:task-descriptions}). 

Our analogies follow the definition given by Structure Mapping Theory (Gentner, 1983), where common relational structures between two domains (i.e, stories, in our setting) define an analogy. Succeeding on our tasks does not involve recalling the surface form of stories, but rather pin-pointing the shared relational structures. Longer stories preserve the relational structures but are padded with “noise.” When humans perform our task, we intend for them to come up with their own internal representation of salient features. Our benchmark then focuses on the question of how LLMs fare when they are presented with the same task.

\subsection{Design Considerations and Challenges} 
\label{subsec:desiderata}

Benchmarks come with design principles and necessary assumptions. We discuss the unique qualities of analogical reasoning that guide and motivate our design and lay out important assumptions in our benchmark. 

\vspace{-0.5mm}
\paragraph{Assess the breadth of analogies.} 
The universe of analogies is vast, and any LM is likely only able to predict a small (often easy) subset of this universe.
While measuring the precision of LMs is important, an ideal benchmark should also measure their recall (how well they capture deep and abstract analogies). 
Generative evaluation might not fully capture this depth, as there may exist many analogies that the LM cannot predict. 
To assess what an LM cannot predict, we propose a set of analogies of our own choosing, and evaluate analogical identification on this set (\S\ref{sec:task-descriptions}). An LLM which has trouble recognizing analogies would also likely have trouble applying them in diverse and meaningful ways. Since recognizing analogies seems to be a bottleneck, we focus our research towards this area. 

\vspace{-0.05cm}

\paragraph{Benchmark size and diversity} The purpose of our dataset is to probe the extent of analogical ability in LLMs, which we are able to show is somewhat limited. Our purpose is \textit{not} to create a set of analogies that covers the universe of possible analogies, but rather to propose specific cases that challenge an LLM’s capability. For example, it would not be useful to construct a broad set of simple analogies which all considered LLMs can trivially solve. We thus design our benchmark to explore the limitations of current LLMs in their analogical reasoning ability. 

\paragraph{Task objectivity}
The quality of real world analogies inherently lie on a spectrum---some are stronger and some are weaker \citep{gentner1983structure}. Ideally, a measure of analogical reasoning encompasses both stronger and weaker analogies. 
Our task aims to capture the inherent subjectivity of analogies while remaining fundamentally \textit{objective}. 
We frame our analogical identification task more specifically as a ranking task, where the best answer must be preferred over evidently incorrect choices.
In doing so, we can measure performance on analogies of differing strength, while maintaining objectivity
(\S\ref{subsec:gold-labels}).

\paragraph{Creativity of analogies.}

LMs perform worse on creative (i.e. rare) data \citep{kandpal2023large}. Thus a benchmark that aims to challenge LMs should feature analogies that are creative. To that end, we introduce novel and diverse, human-written analogies created through a semi-randomized process~(\S\ref{subsec:creation}).

\begin{figure*}[ht]
    \centering
    \includegraphics[scale=.68,trim=4.7cm 5.8cm 6cm 3cm]{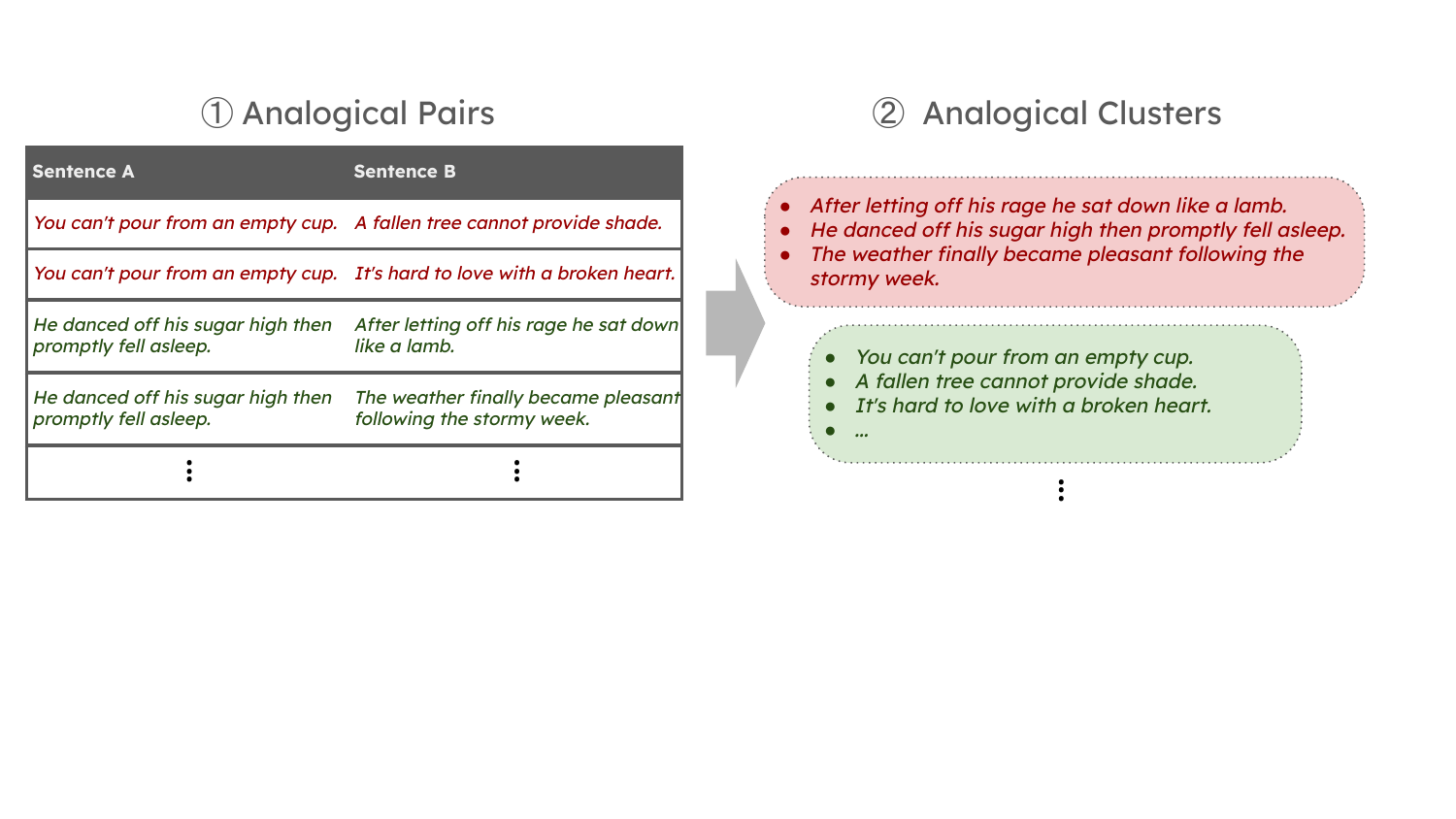}
    \caption{An overview of dataset creation (\S\ref{subsec:creation}). \textbf{\circled{1} Left:} Human annotators are asked to create pairs of analogous sentences. Sentences can be repeated from analogy to analogy. \textbf{\circled{2} Right:} Pairs that share a sentence can be grouped into a cluster of mutually analogous sentences by transitivity.
    }
    \label{fig:dataset}
\end{figure*}

\subsection{Dataset Creation} 
\label{subsec:creation}

\paragraph{\circled{1} Curating analogical sentence-pairs.}
We collected 340 analogies from 4 human annotators (the authors) after multiple rounds of editing. The human annotators included native English speakers and non-native speakers who all attended university in the United States. 
These analogies then served as true positives in our experiments. 
We prioritized quality over quantity, as initial attempts to collect data using a large pool of crowdworkers (AMT and Prolific) yielded low quality annotations. Since our benchmark places high importance on the quality of analogies (\autoref{subsec:desiderata}), we opted to use our current annotation scheme instead.
These annotations of analogies are arranged in pairs of sentences that share similar relational patterns.
For example, in Fig.~\ref{fig:dataset} the sentences ``He danced off his sugar high then promptly fell asleep'' and ``The weather finally became pleasant following the stormy week'' form an analogy. While these two sentences are topically dissimilar (dancing vs weather), they nevertheless share abstract relational patterns.

The construction of these sentence pairs follows this process: For each annotation, a random sentence is provided to the annotator, who is tasked with creating a corresponding analogous sentence. 
Source sentences were sampled from Cambridge Dictionary examples of idioms found on an online resource\footnote{See this 
\href{https://web.archive.org/web/20240409012332/https://www.ef.edu/english-resources/english-idioms/}{link}
.} and a dataset of metaphors~\cite{bizzoni-lappin-2018-predicting}  filtered down to keep only examples with the strongest or second-strongest grades.
To encourage innovative and abstract analogies, the annotator is given 3 random words to incorporate in the newly formed sentence.\footnote{Randomization was achieved by sampling nouns, verbs, and adjectives from 
\href{https://web.archive.org/web/20231209010137/https://www.mit.edu/~ecprice/wordlist.10000}{here}.
} 
During our pilot study, the introduction of random words was found to induce more creative annotations.

There are 
guidelines that the annotators adhere to. Firstly, they are instructed to avoid using similar topics or words as the original sentence. 
This is to eliminate any easy shortcuts that might allow LMs to recognize an analogy without having identified relational patterns. For example, an LM might mistakenly use similar phrasing between a pair of sentences to detect an analogy.  
Instead, the analogy between two sentences should be established on the basis of shared relational patterns. 
Finally, a separate reviewer scrutinizes the contributed sentence pairs to ensure their clarity, accuracy, and effective use of analogy.

\paragraph{\circled{2} Forming analogical clusters.}
Our collected data is structured such that the same sentence can appear in several analogous sentence pairs.
This allows us to organize our dataset into sets of analogous \textit{clusters}, where all pairs of sentences in a cluster are mutually analogous by transitivity.
Each cluster is manually inspected and adjusted to confirm mutual analogousness.
Furthermore, different clusters that happen to share common relational structures are combined.
We then use these clusters to setup the tasks in \S\ref{sec:task-descriptions}.

\paragraph{\circled{3} Analogy elaboration.} \label{sec:elaboration}

To investigate the effect of story length on the complexity of analogies, we collect elaborated versions of each story.
First, longer stories requires analogical reasoning over longer contexts, a task which scales in difficulty for LMs, as shown by the recent results~\cite{chen2023extending,liu2023lost}. 
Second, the longer the stories in an analogy, the more room for expressing abstract relational patterns. 

To implement this elaboration, we use GPT-4 
to convert sentence-level analogies into detailed stories with a target length of 10 sentences and 30 sentences. We selected GPT-4 for its advanced story generation capabilities and proficiency over other LMs in generating coherent and complex text.\footnote{In our pilot experiments, we compared the elaborations using GPT-4, PaLM and Claude, and ultimately chose GPT-4 because of its accurate yet creative elaborations.} To balance creativity and logical consistency, we configured the model with parameters such as temperature $=1$ and top-$p=0.95$. We provide the prompt templates used in Appendix~\ref{appendix:expansion}. Although we later evaluate GPT-4 on its own generations, we demonstrate that self-evaluation bias does not affect our conclusions by testing GPT-4 on stories generated by a different model (\S\ref{subsec:self-generated}).

\paragraph{Statistics.}
\autoref{tab:dataset-summary} shows the overall statistics of our collected analogical clusters and their elaborations. 
Overall, we compiled a total of 340 stories grouped into 47 clusters. On average, each cluster consists of about 7.2 stories. 

\begin{table}[ht]
    \centering
    \small
    \resizebox{0.94\linewidth}{!}{
    \begin{tabular}{lr}
       \toprule 
       Measure & Value \\ 
       \midrule
       \# of clusters & 47.0 \\
       avg. size of clusters (stories) & 7.2 \\
       avg. length (sentences) of 1 sentence stories & 1.2 \\
       avg. length (sentences) of 10 sentence stories & 11.9 \\
       avg. length (sentences) of 30 sentence stories & 31.2 \\
       avg. length (subwords) of 1 sentence stories & 21.3 \\
       avg. length (subwords) of 10 sentence stories & 225.8 \\
       avg. length (subwords) of 30 sentence stories & 552.8 \\
       \bottomrule
    \end{tabular}
    }
    \caption{Summary of dataset statistics. The dataset consists of 47 clusters with an average of 7.2 stories each, and stories vary in average sentence and subword length.}
    \label{tab:dataset-summary}
\end{table}

\subsection{Analogy Identification Tasks}
\label{sec:task-descriptions}
\label{subsec:t2}
\label{subsec:t1}

With the clusters of analogies defined (\S\ref{subsec:creation}), we leverage this data to devise two tasks to benchmark the capability of state of the art LMs at analogical reasoning. In \autoref{sec:intro}, we introduced two components of analogy making. Each task aims to evaluate both components in conjunction. 
Given a query story, both tasks involve identifying analogous stories to the given one from a story bank. 
In the first task, we maintain a small story bank to focus the challenge on rating a few candidates, thereby disentangling it from the challenge of long-context reasoning. 
In the second task, given a story, a model must identify analogous stories from a large story bank.
We intend this approach to be analogous to how humans recollect and employ their past experience to form analogies.

\paragraph{\emph{\textbf{T}}$_1$: Identify analogies from mini story bank.}  
This task confronts the model with choosing the most fitting analogy from 4 options. Given a sentence or story, the model must select the most suitable option from a lineup consisting of one correct answer and three distractors to assess discernment of analogical relationships. 
Negative examples are identified with the help of the analogy clusters identified in \autoref{subsec:creation}. By construction, for a given story, all stories within its cluster are analogous, and all stories in the complement are not. Thus, negative examples are sampled from the complement.
Each answer choice is prefixed by a letter from [A, B, C, D] (e.g. ``D. A fallen tree cannot provide shade''). We prompt each model to answer the question: ``Which of the following is the most analogous story to the target story?'' To guide the LM to make a selection, we impose an additional condition in the prompt that the generation must be one of the four letters. More details of our approach can be found in \autoref{sec:appendix-}.

An example of this task is shown in Fig.~\ref{fig:intro-figure}.
Note, given the elaborated 
stories  discussed in (step \circled{3} in \autoref{subsec:creation}), we have three datasets of multiple-choice questions for each story length (1-sentence, 10-sentences, 30-sentences).

\paragraph{\emph{\textbf{T}}$_2$: Identify analogies from large story bank.}

In this task, given a story, the model must identify the top 10 most analogous stories from a carefully assembled, fixed story-bank consisting of 200 different stories.
This task can be thought of as an extension of the previous task, where there are 200 candidates instead of 4. 
Each story is prefixed by a number from 1 to 200 (e.g. ``1. Kim checked the papers...'').
For this task, we prompt each model to generate a list of integers representing the index numbers of its selections. Following this, we employ precision and recall metrics to analyze its performance. More details and examples are provided in \autoref{sec:prompts-full}.

Like the earlier task, we study this task in three distinct setups as a function of story length (1 sentence, 10 sentences, 30 sentences). The size of the story-bank provided to the model varies considerably on different datasets. For 1-sentence dataset, the story-bank for it contains 4K tokens. For 30-sentence story-bank, it contains 110K tokens. 
The long-context nature of this task poses a major challenge for LMs. 
Due to these constraints, our 
evaluation of \TTwo{} (\S\ref{subsec:analogy-results}) is limited to the few models (GPT-4 and \ClaudeVTWo) that can handle long-context. 
Additionally, while human annotation is possible for this task, it would be impractically expensive, and as such were unable to measure human performance on this task.

\section{Main Experiments}
\label{sec:experiments}

We structure our experimental assessment around two primary tasks aimed at evaluating the analogical reasoning of LMs. 
We discuss the experimental setting including the metrics, models and human evaluation (\S\ref{subsec:setting}), followed by the results.

\subsection{Experimental Setting}
\label{subsec:setting}
\paragraph{Metrics.}
All scores are reported as percentages. 
For \TOne{} (analogies from a \emph{small} story bank, \S\ref{subsec:t1}) we use accuracy as the primary measure of success. 
Each example has multiple candidate analogies. A solver gets a score of $1$ if it chooses the most analogous story and
$1/k$ if it reports no-answer or a $k$-way tie that includes the correct answer ($k=4$ in our dataset.) For \TTwo{} (analogy from a \emph{large} story bank, \S\ref{subsec:t2}), we report common retrieval metrics such as Mean Average Precision (MAP), Precision@K, Recall@K, and Mean Reciprocal Rank (MRR) ~\cite{manning2008introduction}. 

\paragraph{Evaluated models.}
We include models of varying sizes and architectures in our benchmarks. The models include \GPTFour~\cite{openai2023gpt4}, \GPTThreeFive~\cite{brown2020language}, \LLaMATwo-chat~\cite{touvron2023llama2}, , \XwinLM~\cite{xwin-lm}, \WizardLM~\cite{xu2023wizardlm}, \TuluTwo~\cite{ivison2023camels}, \Zephyr~\cite{tunstall2023zephyr}, \ClaudeVTWo~\cite{claude2}, as well as text-to-text models such as \UnifiedQA~\cite{khashabi2020unifiedqa,khashabi2022unifiedqa}. To minimize variations in model responses, we set the decoding parameters to temperature $=0.3$ and top-$p=0.95$. 

\paragraph{Human evaluation.}
We conducted human evaluation to measure whether the task is well-defined and has a reasonable quality. 
This process was meticulously applied to our \TOne{} task (Analogy Selection, \S\ref{subsec:t1}) across different levels of complexity: 1-sentence, 10-sentence, and 30-sentence scenarios. To make the 30-sentence task more manageable, the annotations were done for 30 instances. 
However, for 1-sentence and 10-sentence settings, we annotated 50 instances. 

For each level of complexity, we enlisted three additional annotators\footnote{These annotators share the same demographic as the other annotators in \S\ref{subsec:creation} and were not aware of the experimental design during annotation} (who were not involved in the dataset construction) to evaluate the analogy scenarios. Each annotator began by selecting their personal answer choice without conferring. 
This exercise led to high-agreement among the annotators (\S\ref{subsec:gold-labels}).  

Following this individual judgment phase, disagreements were adjudicated via discussion among the annotators. During these discussions, the annotators were encouraged to exchange their rationales behind their initial selection and converge upon one collective answer that we used for evaluation. 

We did not run human annotations for \TTwo{} due to the immense reading load expected of annotators. 
Since the two tasks are based on the same set of labeled data, we focus our human annotations on \TOne{} to establish the quality of the presented data.

\subsection{Result: Mini Story Bank (\TOne)}
\label{sec:results}

We benchmark how well our models can identify analogies from a mini story bank (so as to disentangle this task from other challenges associated with long-context reasoning). Our results are reported in \autoref{tab:models_performance} and Fig.~\ref{fig:combined}.
More detailed results are reported in \autoref{tab:models_all_performance} of the Appendix. 
Overall, our analogical reasoning benchmark challenges state of the art language models.

\paragraph{LMs do not outperform humans.}
The results reveal that analogical ability varies widely among modern LMs. 
While many models perform non-trivially (i.e. better than 25\% accuracy achieved by random guessing), and some models such as \GPTFour~perform considerably well, no model is able to outperform humans in any setting.  
Among open-source models, the largest models (70B) dominate the results for the shortest story length setting, with the exception of UnifiedQA which is supervised with different data than the rest of the models.

\begin{table}[ht]
    \scriptsize
    \setlength{\tabcolsep}{2pt} 
    \small
    \resizebox{0.99\linewidth}{!}{
    \begin{tabular}{clccc}
       \cmidrule[1pt]{2-5}
       & Model $\shortdownarrow$ - Story length $\shortrightarrow$ & 1-sent & 10-sent & 30-sent \\ 
       \cmidrule[0.5pt]{2-5}
        & \textcolor{gray}{Random} & \textcolor{gray}{25} & 
       \textcolor{gray}{25} & \textcolor{gray}{25}  \\
        \cdashline{2-5}\noalign{\vskip 0.5ex}
       \multirow{6}{*}{\rotatebox{90}{{Open-source}}} & Zephyr (7B) & 55.1 & 27.1 & 20.3 \\
       & UnifiedQA (11B) & 68.1 & 27.3 & 17.8 \\
       & WizardLM (13B) & 41.1	& 29.1 & 25.7  \\
       & \LLaMATwo-chat (70B) & 55.6 & 39.2 & 29.5  \\
       & XwinLM (70B) & 66.3 & 35.7 & 26.8 \\
       & Tulu2 (70B) & \textbf{71.8} & \textbf{51.2} & \textbf{31.5}  \\
       \cdashline{2-5}\noalign{\vskip 0.5ex}
       \multirow{3}{*}{\rotatebox{90}{{Private}}} 
       & Claude & 68.2 & 30.2 & 25.9 \\
       & GPT3.5 & 65.3 & 46.4 & 30.8 \\
       & GPT4 & \textbf{89.1} & \textbf{66.5} & \textbf{60.7}  \\
       \cdashline{2-5}
       \noalign{\vskip 0.2ex} 
       & \cellcolor{intgray}  \textbf{Human} & \cellcolor{intgray} \textit{\textbf{96.0}} & \cellcolor{intgray} \textit{\textbf{72.5}} & \cellcolor{intgray} \textit{\textbf{73.3}} \\
       \cmidrule[0.8pt]{2-5}
    \end{tabular}
    }
    \caption{Benchmarking various models for \TOne{} (\S\ref{sec:results}). 
    For open-source models, we only show the results of the largest available sizes in their model family. 
    \textbf{While the best models perform somewhat close to human in short analogies (1-sentence), the human-AI gap increases in longer stories}. 
    }
    \label{tab:models_performance}
\end{table}

\begin{figure*}[ht]
    \centering

    \includegraphics[width=0.84\textwidth,trim=0cm 0cm 0cm 0.0cm]{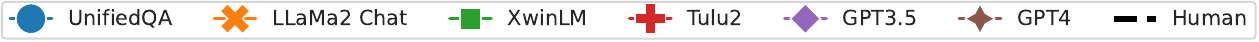}
    \begin{minipage}[b]{0.67\textwidth}
        \centering
        \subfloat[Results with varying model scale. 
        The error margins are based on the standard error. 
        While scaling LMs is effective among short (1-sent) stories (left),  
        \textbf{the benefit of scale is negligible for longer stories} (middle and right).]{
            \includegraphics[width=\textwidth,trim=0cm 0.1cm 0cm 0.0cm]{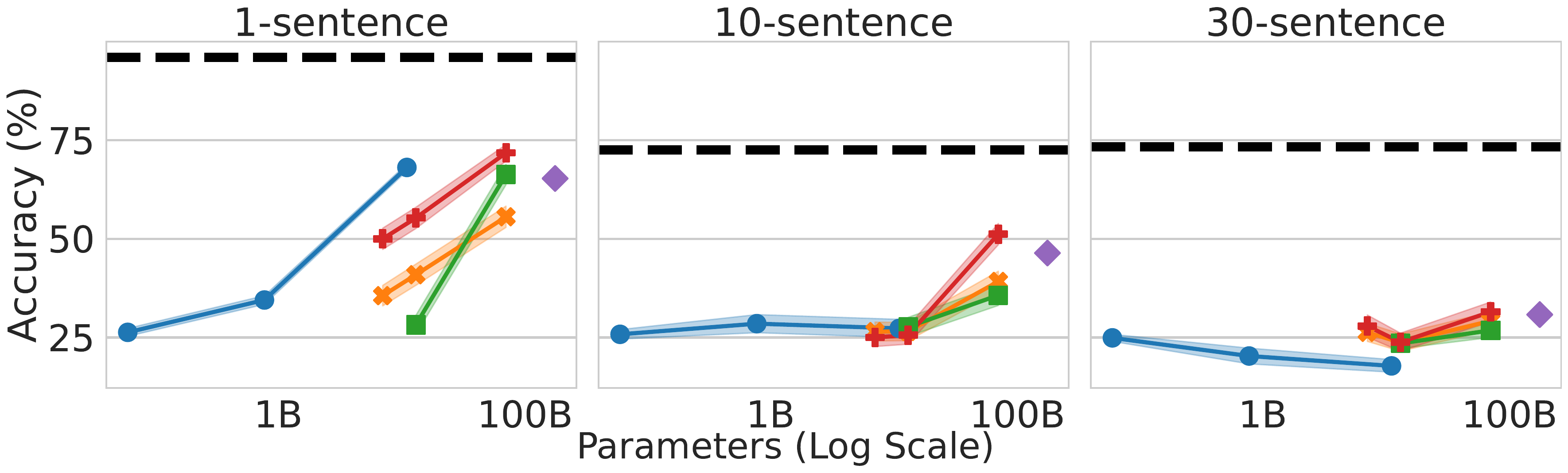}
            \label{fig:scale}
        }
    \end{minipage}
    \hspace{.4cm}
    \begin{minipage}[b]{0.25\textwidth}
        \centering
        \subfloat[With increasing story length, 
   \textbf{model accuracy decreases, while their gap with humans increases}. ]{
            \includegraphics[width=\textwidth, trim=0cm 0.0cm 0cm 0.0cm]{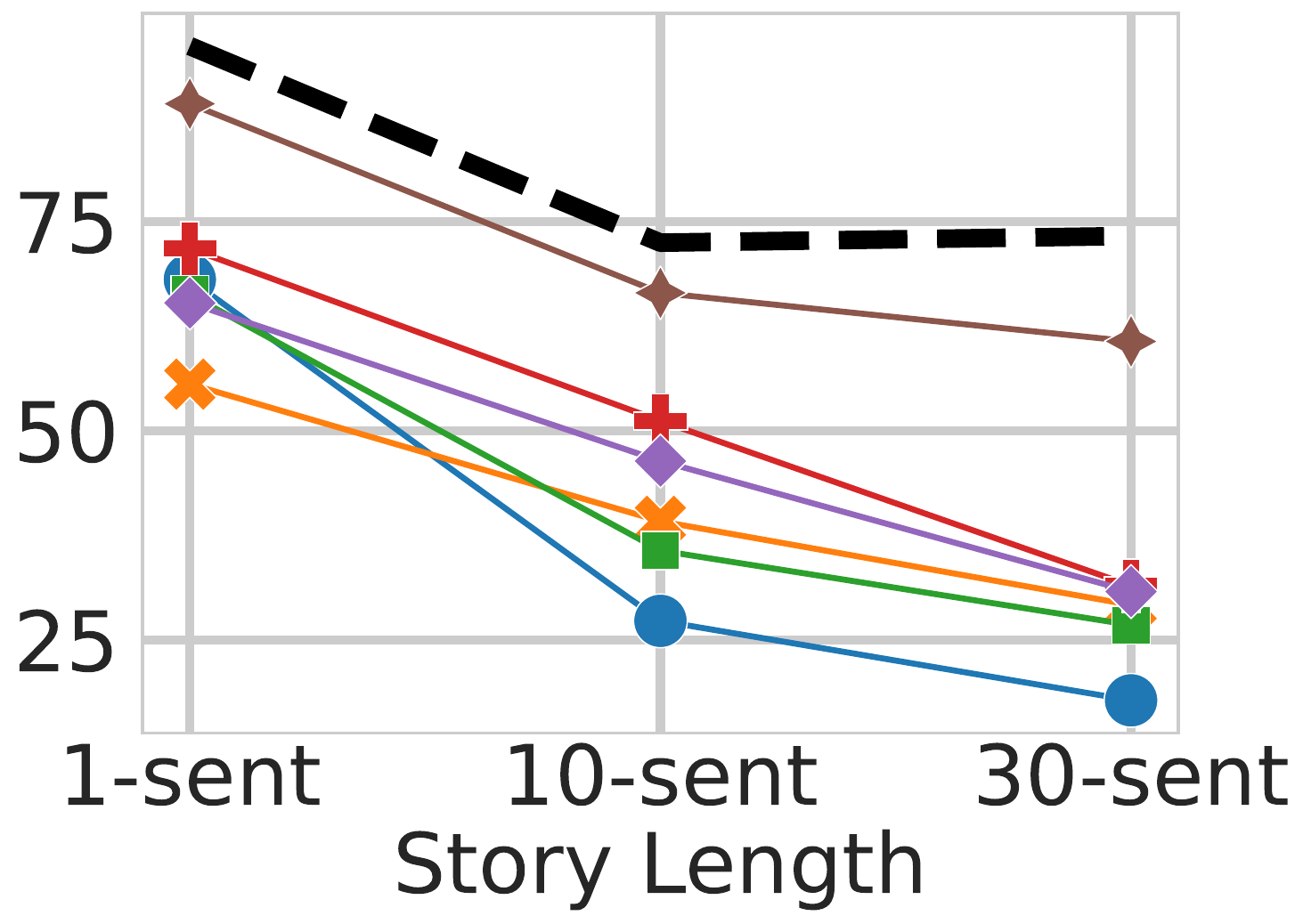}
            \label{fig:sent-length}
        }
    \end{minipage}
   \caption{
   Accuracy of LMs on \TOne{} (\S\ref{sec:results}). 
   }
   \label{fig:combined}
\end{figure*}

\paragraph{Analogy length degrades LM performance.}

We evaluate our lineup of models on stories consisting of 1, 10, and 30 sentences. In Fig.~\ref{fig:sent-length} all models exhibit  degradation as story length increases. In contrast, while human performance also decreases with longer length, their performance stops decreasing for 10- and 30-sentence stories. Thus, the performance gap between humans and LMs increases with longer context-length. These results to suggest that analogical reasoning over longer inputs poses an inherent challenge for LMs. 

\paragraph{Model scaling benefits are limited on long stories.} \label{subsec:scaling-results}
Even if performance diminishes with increased story length across all models, as long as performance improves with model size, a sufficiently large model can solve this problem.
To test this possibility, we evaluate models of varying sizes within the \UnifiedQA, \LLaMATwo, \XwinLM, and \TuluTwo~families on \TOne.
Our results in Fig.~\ref{fig:scale} indicate that while performance scales with LLM size in the single sentence setting, we do not observe the same trend in longer settings. 
Specifically, in longer stories performance plateaus across model family as model size increases.
The observed trend indicates a limit to the benefits of scaling model size when handling complex analogies.

\subsection{Results: Large Story Bank (\TTwo)}
\label{subsec:analogy-results}

Having evaluated our lineup of models on the mini story-bank setting, we now turn our attention to the full story-bank setting. As stated in \S\ref{sec:task-descriptions}, fitting the full story bank in the context window requires us to consider only long-context models such as GPT4 and Claude and precludes human annotation due to the large workload (and corresponding monetary cost) that the task entails. In this experiment, given a story, each model must identify the top $k$ most analogous stories from the story-bank. We report the precision-recall curves for $k=1,...,10$ in Fig.~\ref{fig:precision-recall} and provide further details in \autoref{tab:metrics} of the Appendix.

\paragraph{LM performance approaches random.}
We evaluate both models as well as a trivial baseline where $k$ random documents are retrieved. 
Both models perform similarly to the trivial baseline in most cases. 
An exception is the performance of GPT4-Turbo in the single-sentence setting, suggesting that the task, though challenging, is not impossible for LMs to perform. 
While impressive, the performance of GPT4-Turbo is nevertheless near trivial in lengthier settings. 
These evaluations test the limits of the best modern LMs. 
If humans can recollect relevant experiences to form analogies~\citep{keane1987retrieving, wharton1994below}, then 
our results suggest that further research is necessary to achieve parity in LMs. 

\begin{figure*}
    \centering
    \includegraphics[width=\textwidth, trim={0 4.5mm 0 0}, clip]{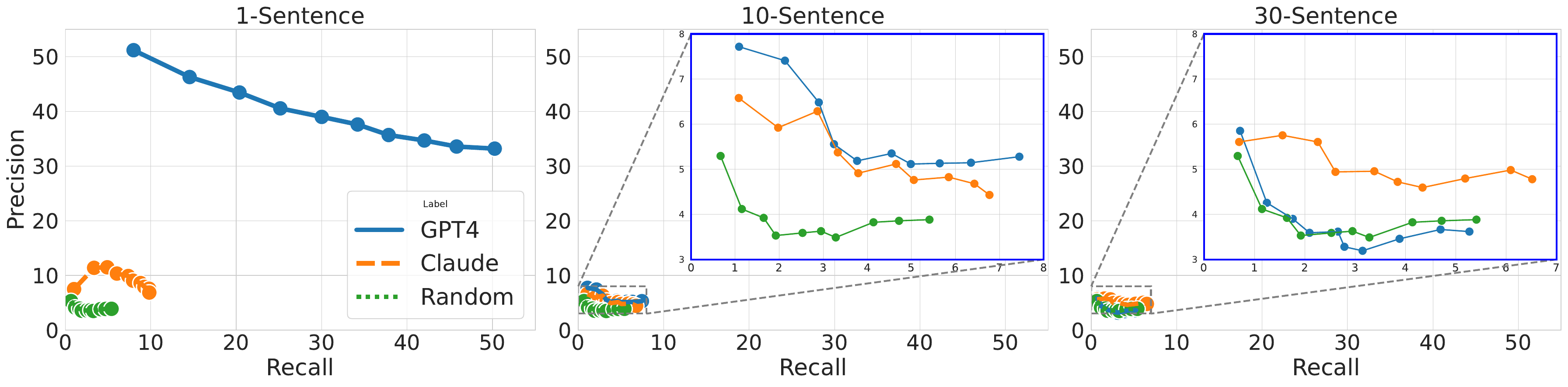}
    \caption{
    Precision-recall plot (in percentage) of LMs on \TTwo{} (\S\ref{subsec:analogy-results}) at three different story lengths (1, 10, 30 sentences).
    \textbf{With increasing story length, the precision-recall of the models approaches random.} 
    }
    \label{fig:precision-recall}
\end{figure*}

\section{Further Analysis}

\subsection{Evaluating Self-Generated Stories}
\label{subsec:self-generated}
In past experiments we utilized GPT-4 to extend single-sentence stories into versions containing 10 or 30 sentences. Consequently, the relatively high accuracy of \GPTFour~may stem from evaluating its own generated content. 
To address this, we also evaluate \GPTFour~on stories generated by Claude. 
As a baseline, we also evaluate  Claude on its own stories and stories generated by \GPTFour. We report our results in \autoref{tab:evaluator_generator} and find that \GPTFour~encounters negligible performance degradation upon switching to Claude generations. Additionally, \GPTFour~consistently outperforms Claude on Claude generations. These results suggest that the relatively high performance of \GPTFour~is likely attributed to factors other than evaluating its own generations. 

\begin{table}[ht]
    \centering
    \small
    \begin{tabular}{ccccc}
       \toprule 
       \multirow{2}{*}{\diagbox{Eval.}{Gen.}} & \multicolumn{2}{c}{10 Sentences} & \multicolumn{2}{c}{30 Sentences} \\ 
       \cmidrule(lr){2-3} \cmidrule(lr){4-5}
       & Claude & GPT4 & Claude & GPT4 \\ 
       \midrule
       Claude & 36.5 & 30.2 & 33.5 & 25.9 \\
       GPT4 & 69.7 & 66.5 & 57.6 & 60.7 \\
       \bottomrule
    \end{tabular}
    \caption{Perf. of different evaluators and generators on 10- and 30-sentence stories (\S\ref{subsec:self-generated}). \textbf{\GPTFour{} performance experiences minimal change when evaluating Claude generations.}}
    \label{tab:evaluator_generator}
\end{table}

\subsection{Effect of Dataset Error}
\label{subsec:gold-labels}
Whether incorrect LM predictions are attributable to dataset error/subjectivity is unclear. 
To reduce the likelihood of dataset error affecting our conclusions, we deem analogies that were correctly predicted by humans (in our human evaluation) to be relatively free of error, and repeat experiment \TOne~on those analogies. 
As \autoref{tab:model_classification_accuracy_human} shows, 
all trends reported in \S\ref{subsec:analogy-results} are still observed in this low-error setting, suggesting that our conclusions are unlikely to be affected by marginal dataset error.

\begin{table}[ht]
    \centering
    \small
    \begin{tabular}{cccc}
       \toprule 
       Model & \multicolumn{3}{c}{\TOne: Classification (accuracy\%)} \\
       \cmidrule{2-4}
       & 1 sentence & 10 sentences & 30 sentences \\
       \midrule
       GPT4 & 91.7 & 58.3 & 22.7 \\
       Tulu7 & 47.2 & 30.1 & 16.7 \\
       Tulu13 & 52.8 & 30.1 & 22.2\\
       Tulu70 & 74.0 & 34.7 & 25.0 \\
       Xwin13 & 27.8 & 19.4 & 22.9 \\
       Xwin70 & 42.2 & 31.9 & 29.1 \\
       \bottomrule
    \end{tabular}
    \caption{Model accuracy on true positive human predictions in \TOne~(\S\ref{subsec:gold-labels}) at three different story lengths (1, 10, 30 sentences). \textbf{All trends are consistent with the original task.}}
    \label{tab:model_classification_accuracy_human}
\end{table}

\paragraph{Agreement among human annotators.}
Using our task definition, when we measure the inter-annotator agreement on our human-written analogies (1-sentence), we find that all three human evaluators agree unanimously on 47 of 50 analogies. The high degree of inter-annotator agreement is a quantitative indicator of our dataset's objective evaluation and quality. In the 10- and 30-sentence settings, agreement decreases to $0.70\%$ and $0.73\%$ of analogies respectively. Given the quality of the extended stories as attested in \autoref{appendix:expansion}, we attribute this decrease in agreement to the increased difficulty of these settings.

\subsection{Longer Analogies are Easier for Humans}
\vspace{-0.02cm}

In \autoref{subsec:creation} we hypothesized that analogy length corresponds to complexity. While our results clearly indicate that longer analogies pose a greater challenge for LMs, perhaps a more interesting question is whether they pose a greater challenge for humans. Surprisingly, qualitative feedback from human annotators indicated that they found the 30 sentence setting easier than the 10 sentence setting, observing that added details in the longer setting aid in disambiguation when performing the task. While we expected annotator performance and agreement to \textit{decrease} in the longest setting, we \textit{did not} observe this trend in either result (\autoref{tab:models_performance}, \autoref{subsec:gold-labels}).

\vspace{-0.06cm}
\section{Discussion}

\vspace{-0.06cm}
\paragraph{Limits of modern LMs in analogical thinking.}
A clear consensus on whether LMs can adequately perform analogical thinking has remained elusive. 
While some find that LMs are proficient analogical reasoners \citep{webb2023emergent, ichien2023large}, others have challenged this notion \citep{jiayang2023storyanalogy}. 
Throughout our experiments, we repeatedly find that modern LMs display limited ability to engage in key aspects of analogical thinking.
Crucially, performance does not scale with model size on longer stories.
Unlike the LMs evaluated, humans can identify analogies between even the longest stories with relative ease.
These observations clearly suggest that LMs lack some key mechanism to think analogically.
Overall, our results establish the need for further research to encourage analogical thinking in LMs. 

\vspace{-0.09cm}
\paragraph{Downstream applications and future work.}
\label{subsec:future-work}
What downstream applications can we expect from analogically reasoning LMs?
We discuss examples to illustrate the potential of analogical LMs. In science, analogy provides a source of inspiration for innovation. For instance, the design of artificial neural networks was inspired by biological neural networks \citep{rosenblatt1958perceptron}. An analogy driven scientific search engine would accelerate such innovation, allowing researchers to consider relevant ideas across vastly different contexts \citep{hope2017accelerating}. In law, \citet{zou2024reframingtaxlaw} has represented consistency in legal decisions as an analogical reasoning problem, where decisions in a current case should follow that of an analogous case. An analogical search engine would aid in the identification of relevant cases. 
Given these wide-ranging applications, we hope that our findings motivate future work towards equipping LMs with better analogical reasoning capabilities.

\vspace{-0.06cm}
\section{Conclusion}
\vspace{-0.08cm}
Analogical reasoning is an important aspect of human cognition, with wide-ranging potential for future research.
To  benchmark this ability in LMs, we define a general approach by scaling the length of stories and the context from which they need to be retrieved. Our benchmark exposes the limitations of analogical reasoning in modern LMs.  We release \name{} to motivate further research.

\section*{Limitations}

In our experiments we benchmark many models. While trying more models and performing additional prompt-engineering could have affected results, in the end we were constrained by the available computing resources.
Additionally, we cannot exclude the possibility that LMs encountered labelled analogies during training or finetuning, especially proprietary models such as \GPTFour. While our dataset is more challenging than existing ones, it comes with various simplifying assumptions and cannot capture the potentially-infinite range of analogies.
Future work should extend the existing datasets to capture more complex forms of analogical reasoning and experiment with different prompting strategies.

\paragraph{Analogical reasoning w/ parametric knowledge.} 
Pretraining provides LMs with ample parametric knowledge \citep{brown2020language}, which may be leveraged for analogical reasoning \citep{yasunaga2023large}. 
Our benchmark does not evaluate this ability in LMs, as it would make the evaluation of analogical reasoning difficult to conduct in an objective manner. Compared to our current approach, which controls exactly what stories an LLM has access to, the information stored in the parameters of a network is less certain. Properties such as the difficulty of a question/example are greatly affected by the LLM’s knowledge, which we cannot ascertain. To make our benchmark more objective, we leave the evaluation of parametric knowledge to future work, and focus our research on retrieving stories in-context. Our results nevertheless yield valuable insight on the limitations of LLMs. 

\section*{Ethical Considerations}
We hereby acknowledge that all authors of this work are aware of the provided ACL Code of Ethics
and honor the code of conduct.
The work presented here does not immediately raise any ethical concerns, to our knowledge.  
Beyond the scope of this work, analogical reasoning should be applied with care, otherwise due to its inherent subjectivity
it may potentially lead to misleading or incorrect conclusions.

\section*{Acknowledgements}
This work is supported by a generous gift the Allen Institute for AI and partly by ONR grant (N00014-24-1-2089). 
We are grateful to Yejin Choi, Ben Van Durme, Candice Penelton, Jack Zhang and Jiefu Ou  for their insightful feedback throughout this project. 
GPU machines for conducting experiments were provided by ARCH Rockfish cluster (\url{https://www.arch.jhu.edu}).

\bibliography{ref}
\bibliographystyle{acl_natbib}

\clearpage
\appendix
\onecolumn
\begin{center}
{\Large \textbf{Supplemental Material}}
\end{center}

\section{Additional Related Work}
Here we cover additional related work that did not fit in the main text. 

\paragraph{Analogical reasoning before LMs.}
The research on analogical reasoning in AI and cognitive science for the longest time has focused on four-term analogies~\citep{hesse1965models} (e.g., ``Baltimore to Maryland is like NYC to New York''). 
In the era of symbolic AI era, 
an extensive literature focused on engineering symbolic systems that processed analogical reasoning~\citep{winston1980learning,carbonell1983learning,hofstadter1984copycat,schank1999dynamic}. These works focus on richer representation for alignment of analogous symbols and their dynamic retrieval from a memory structure.

The more complex the analogies are, the more complex representation they require~\cite{holyoak2001place}. Naturally, it meant that solving the analogy problem require solving the representation problem. 
The increasing progress in extracting representations of language led to more progress in analogical reseasoning. 
A decade ago, the earlier generation of representation learning algorithms such as Word2Vec ~\citep{mikolov2010recurrent,mikolov2013efficient} famously showed linguistic regularities equivalent to 
lexical analogies~\citep{pennington2014glove,ethayarajh2018towards}
Thereafter, a large body of works focused on effective ways of eliciting analogies from word embeddings~\citep{murena2020solving}, sometimes through neural networks or symbolic reasoning frameworks built atop these embeddings~\citep{lamm2018textual,alsaidi2021neural,marquer2023solving}.

\paragraph{Analogical reasoning in humans.}
The cognitive ability to process analogies likely has been with homosapiens since the time they developed their languages, as evidenced by written Babylonian or Egyptian relics~\citep{holyoak1996mental}.
These written documents convey a variety of ideas: friendship and emotions, dangers and enemies, power and greed, and so on. 

Analogies also made their way to science. Greeks used analogies to describe their understanding of physical concepts, such as sound waves spreading like water waves. Physicists used similar abstractions to understand light waves by formulating analogies to known physical waves, leading to ``wave theory of light''.  
Analogies are so prevalent in scientific development that renowned physicist J. Robert Oppenheimer called it an ``indispensable and inevitable tool for scientific progress'' \citep{oppenheimer1956analogy}.

Cognitive science is the community which adopted a scientific and systematic treatment of analogical reasoning in human cognition.   
Within cognitive science, analogical reasoning was viewed as mental models that utilize structure alignment via relations \citep{gentner1983structure,clement1991systematicity}.
Analogical reasoning was also studied under pragmatic contexts such as the goal of the environment or the problem solving~\citep{gick1980analogical}.
\citet{hofstadter2001analogy,gentner2017analogy} argue that analogical reasoning is the ``core of cognition''.

\clearpage

\section{Examples of seed analogies} \label{appendix:seed-analogies}
We compare seed story examples below between our dataset and \storyanalogy. Each row represents a pair of analogous stories. We find that the analogies in StoryAnalogy share syntactic/surface patterns between stories, which we propose may act as shortcut features in the task of analogy identification.

\begin{table}[ht]
\centering
\begin{tabular}{p{.4\linewidth}p{.4\linewidth}} 
 \toprule
 \textbf{Story 1} & \textbf{Story 2} \\ 
 \midrule
 \cellcolor{intgray}The life of a celebrity looks like they have it all, but in fact they have more problems than anyone can possibly imagine. &	\cellcolor{intgray}Don't trust everything on the social media. It appears that people are having the best time of their lives, but remember, it can be fake. \\ 
  	It is not that cold today, but I'd still go by car since I can't afford to get sick. &  To avoid burning your hands, use oven mitts when removing the cake, as it will be hot unlike how it is now. \\ 
\cellcolor{intgray}I am determined to cook the Thanksgiving dinner from scratch this time no matter how difficult it is.	& \cellcolor{intgray} Harry prepared tirelessly for the final race and put in each and every ounce of energy he had into the training.\\
 \bottomrule
\end{tabular}
\caption{Examples from our approach (\name)}
\end{table}

\begin{table}[ht]
\centering
\begin{tabular}{p{.4\linewidth}p{.4\linewidth}} 
 \toprule
 \textbf{Story 1} & \textbf{Story 2} \\ 
 \midrule
 \cellcolor{intgray}Magma rises from deep in the earth. The magma goes into volcanos &	\cellcolor{intgray}Food goes up from the stomach. The food enters the esophagus. \\ 
  	The stream becomes a river. The river continues to flow along the same path for a long time. &  A person grows from a child into an adult. As time passes, the person experiences ongoing growth and maturation. \\ 
\cellcolor{intgray}The plasma membrane encloses the animal cell. It controls the movement of materials into and out of the cell	& \cellcolor{intgray} Security guards monitor the doors of the factory. They manage the entry and exit of personnel to and from the factory\\
 \bottomrule
\end{tabular}
\caption{Examples from \storyanalogy~\citep{jiayang2023storyanalogy}}
\end{table}

\newpage
\section{Further Details on Analogy Elaboration}
\label{appendix:expansion}

We expand a single sentence to craft a story spanning 10 or 30 sentences. This directive applies to both \GPTFour{} in \S\ref{sec:elaboration}. Below is an example: 
\vspace{0.4cm}

\noindent
\textbf{Example prompts provided to \GPTFour{} for story elaboration: }

\begin{tcolorbox}[colback=gray!5!white,colframe=gray!75!black,title=\textbf{Input:}]
\tt

Given the following sentence, expand it into a 10-sentence story.

Sentence: All that glitters is not gold.

\end{tcolorbox}

\begin{tcolorbox}[colback=gray!5!white,colframe=gray!75!black,title=\textbf{Output:}]
\tt

In the small town of Baker's Crest, people lived simple lives, mostly untouched by the allure of wealth and luxury. The quiet town was buzzing one day with the news of a travelling merchant who claimed to sell "real gold" artifacts. The townspeople gathered at the market square, attracted by the glinting objects displayed on his cart. 

The items glittered under the sunlight, their sheen unmatched, making everyone believe they were indeed made of gold. The merchant, a persuasive man, told captivating tales of the artifacts being lost treasures of ancient kings. An air of greed and desire blew over the town, as everyone wanted a piece of this assumed wealth. They traded their hard-earned savings for the glittering items, believing they had struck a fortune. 

It was not until a wise old man from the town examined the artifacts closely and declared them to be mere brass, cleverly polished to resemble gold. All the glitter had indeed fooled them, reinforcing the adage, "All that glitters is not gold." The realization dawned, leaving the townspeople embarrassed, having learned their lesson the hard way.

\end{tcolorbox}

Throughout this creative process, we regulate it with a temperature setting of 1 and a top\_p value of 0.95. We experimented with different temperatures, but these adjustments introduced additional issues. A high temperature caused the narrative to diverge from the core meaning of the original sentence, whereas a low temperature led to repetitive elements which rendered generated stories highly similar due to shared analogous traits.

\paragraph{Assessing the quality of story expansion.}

We conduct an experiment to test the ability of GPT-4 to extend stories while hewing to the original source. If GPT-4 is successful, then the original source (hypothesis) must entail from the extended story (premise). Modern LLMs are understood to be highly performant on the textual entailment task \citep{srivastava2023beyond, suzgun-etal-2023-challenging}. Thus, we use the recently-released Claude-3 to predict entailment, taking care to avoid any potential bias in these evaluations that might unfairly favor the generations of GPT-4. As baselines, we randomly pair the premise and hypothesis for the 10- and 30-sentence setting.

\begin{table}[ht]
\centering
\begin{tabular}{cc} 
 \toprule
 
Story Comparison & Entailment Rate \\ 
 [0.5ex] 
 \midrule
 1 vs 10 sentence (random) &	0.01 \\ 
 1 vs 10 sentence &	0.95 \\ 
 1 vs 30 sentence (random)	& 0.03	 \\
 1 vs 30 sentence	& 0.97 \\ 
 \bottomrule
\end{tabular}
\end{table}

We show that nearly all our source stories entail from the extended versions.

\newpage

\section{Prompts Used for Evaluating LMs for \TOne{} }
\label{sec:appendix-}

\begin{figure*}[ht]
    \centering
    \includegraphics[width=\textwidth,trim=0cm 7cm 0cm 0cm]{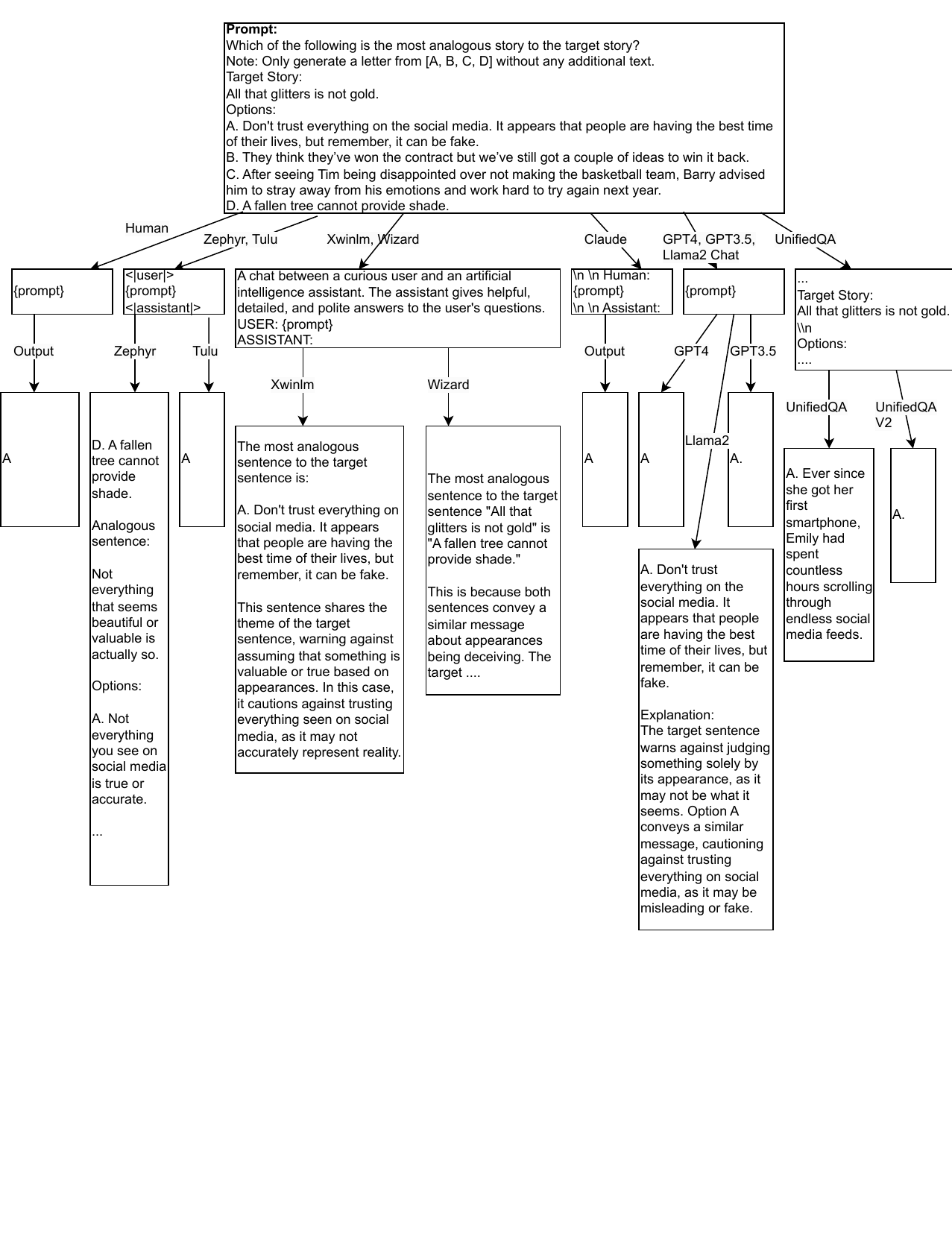}
    \caption{Analogy Selection Prompt for Different Models}
    \label{fig:selection_prompt}
\end{figure*}

Fig.~\ref{fig:selection_prompt} demonstrates the adaptation of a basic prompt to run various model evaluations for \TOne{} task. 
We begin with the basic prompt and adjust it slightly to comply with the specific instructions of each model, as depicted in the second tier of the diagram. The third tier presents examples of responses generated by the models. Also, we set the temperature=0.3 and top\_p=0.95 for all of the model evaluations. 

\newpage
\section{Detailed Results for \TOne{}}

\autoref{tab:models_all_performance} in our research paper presents the comprehensive set of results from our \TOne{} experiments discussed in \S\ref{sec:results}. We assessed the abilities of numerous open-source models as well as \GPTFour{} and \ClaudeVTWo{} on this particular task. We use 4xA100 to evaluate all of the models.

\begin{table*}[ht]
    \centering
    \small
    \begin{tabular}{lccccc}
       \toprule 
       Model & Number of params & \multicolumn{3}{c}{\TOne Analogy Selection (accuracy)} &  \\
       \cmidrule(lr){3-5}
       & & 1 sentence & 10 sentences & 30 sentences \\ 
       \midrule
       Random & -- & 25\% & 25\% & 25\%  \\
       UnifiedQA & 11B & 68.1\% & 27.3\% & 17.8\% \\
       UnifiedQA v2 & 11B & 53.8\% & 29.1\% & 23.6\%  \\
       \LLaMATwo-chat & 7B & 35.6\% & 26.5\% & 26.3\% \\
       \LLaMATwo-chat & 13B & 40.9\% & 26.5\% & 23.7\% \\
       \LLaMATwo-chat & 70B & 55.6\% & 39.2\% & 29.5\% \\
       XwinLM & 13B & 28.2\% & 27.7\% & 23.5\% \\
       XwinLM & 70B & 66.3\% & 35.7\% & 26.8\% \\
       WizardLM & 13B & 41.1\%	& 29.1\% & 25.7\%  \\
       Tulu2 & 7B & 50.0\% & 25.0\% & 27.9\% \\
       Tulu2 & 13B & 55.3\% & 25.6\% & 23.8\% \\
       Tulu2 & 70B & 71.8\% & 51.2\% & 31.5\%  \\
       Zephyr & 7B & 55.1\% & 27.1\% & 20.3\% \\
       GPT3.5 & 175B & 65.3\% & 46.4\% & 30.8\% \\
       GPT4 & ? & 
       \textbf{89.1}\% & \textbf{66.5\%} & \textbf{60.7\%}  \\
       Claude & ? & 68.2\% & 30.2\% & 25.9\% \\
       Human & -- & 96.0\% & 72.5\% & 73.3\% \\
       \bottomrule
    \end{tabular}
    \caption{Performance of different models on analogy selection tasks.}
    \label{tab:models_all_performance}
\end{table*}

\newpage 
\section{Prompts used for evaluating LMs for \TTwo{}}
\label{sec:prompts-full}

In \S\ref{subsec:t2} we discuss \TTwo{}, which is identifying the top 10 most analogous stories from a fixed bank of 200 stories. The following example shows the detail of the prompt.

\subsection*{\GPTFour{} Model Input and Output}

\begin{tcolorbox}[colback=gray!5!white,colframe=gray!75!black,title=\textbf{Input:}]
\tt

Retrieve the top 10 analogous stories from the sentence bank for the following target story:

NOTE: Only generate an index number without any additional text. For example: 1, 2, 3, 4, 5, 6, 7, 8, 9, 10

\textbf{Target Story:}

All that glitters is not gold.

\textbf{Sentence Bank:}

\textbf{1.}Kim checked the papers in a rush so that she can have more free time. But, now she needs to redo them as half of the class complained.

\textbf{2.} Liam lied to get into the school; Lary did not. Liam had a difficult time trying to hide the deception as a result. But unlike Liam, Lary did not have to worry about anything else, so he had a terrific time.

\textbf{3.} I am sorry, but you would now have to present your work before you can go for the vacation.

\textbf{4.} A fallen tree cannot provide shade.

\textbf{5.} He is the winner of three Grammy awards for god's sake! People consider him to be the god of rap.

...

\textbf{197.} It is not that cold today, but I'd still go by car since I can't afford to get sick.

\textbf{198.} I do not want to spoil your mood but I have to babysit my nephew today.

\textbf{199.} Every cigarette you smoked is a threat to your health.

\textbf{200.} He knocked the nail into the wall with a hammer.

\end{tcolorbox}

\begin{tcolorbox}[colback=gray!5!white,colframe=gray!75!black,title=\textbf{Output:}]

\tt
4, 14, 29, 59, 97, 111, 113, 137, 172, 188

\end{tcolorbox}

We also considered using the LM to assign likelihoods to analogous stories, then ranking the entire story-bank by likelihood. However, the extent to which modern LMs are well-calibrated remains unclear, especially in this domain. We conducted preliminary studies that attempted to score the strength of an analogy between two sentences. Scores were wildly inconsistent between runs and different in-context examples, even on low temperature settings. The factors that contribute to the inconsistent behavior remain unclear, and thus we do not define our task in this manner.

\newpage
\section{Detailed Results for \TTwo{}}

\autoref{tab:metrics} in our research paper presents the comprehensive set of results from \TTwo{}. We assessed the abilities of \GPTFour{} Turbo and \ClaudeVTWo{} on this particular task(\S\ref{subsec:analogy-results}). We use 4xA100 to evaluate all of the models. Here are some detailed results of it:

\begin{table*}[ht]
    \centering
    \small
    \begin{tabular}{@{}lcccccccccc@{}}
        \toprule
        Metrics & \multicolumn{3}{c}{\TTwo: [GPT4-turbo]} & \multicolumn{3}{c}{\ClaudeVTWo} & Random & Oracle \\
        \cmidrule(lr){2-4} \cmidrule(lr){5-7}
         & 1 sentence & 10 sentences & 30 sentences & 1 sentence & 10 sentences & 30 sentences & & \\
        \midrule
        P@3 & 42.9\% & 6.5\% & 3.9\% & 10.2\% & 5.3\% & 5.4\% & 3.9\% & 100\% \\
        P@5 & 38.5\% & 5.2\% & 3.6\% & 8.9\% & 4.6\% & 4.7\% & 3.6\% & 100\% \\
        R@3 & 20.1\% & 2.9\% & 1.8\% & 4.3\% & 2.4\% & 2.2\% & 1.6\% & 48.9\% \\
        R@5 & 29.7\% & 3.8\% & 2.7\% & 6.6\% & 3.6\% & 3.2\% & 2.5\% & 81.6\% \\
        MAP & 55.4\% & 14.2\% & 10.8\% & 6.3\% & 1.9\% & 3.4\% & 1.7\% & 100\% \\
        MRR & 64.2\% & 15.6\% & 11.3\% & 18.9\% & 9.8\% & 13.4\% & 11.1\% & 100\% \\
        \bottomrule
    \end{tabular}
    \caption{Performance metrics for \TTwo{} using  and \ClaudeVTWo{} at different sentence lengths.}
    \label{tab:metrics}
\end{table*}

\begin{figure}[ht]
    \centering
    \includegraphics[width=0.5\columnwidth]{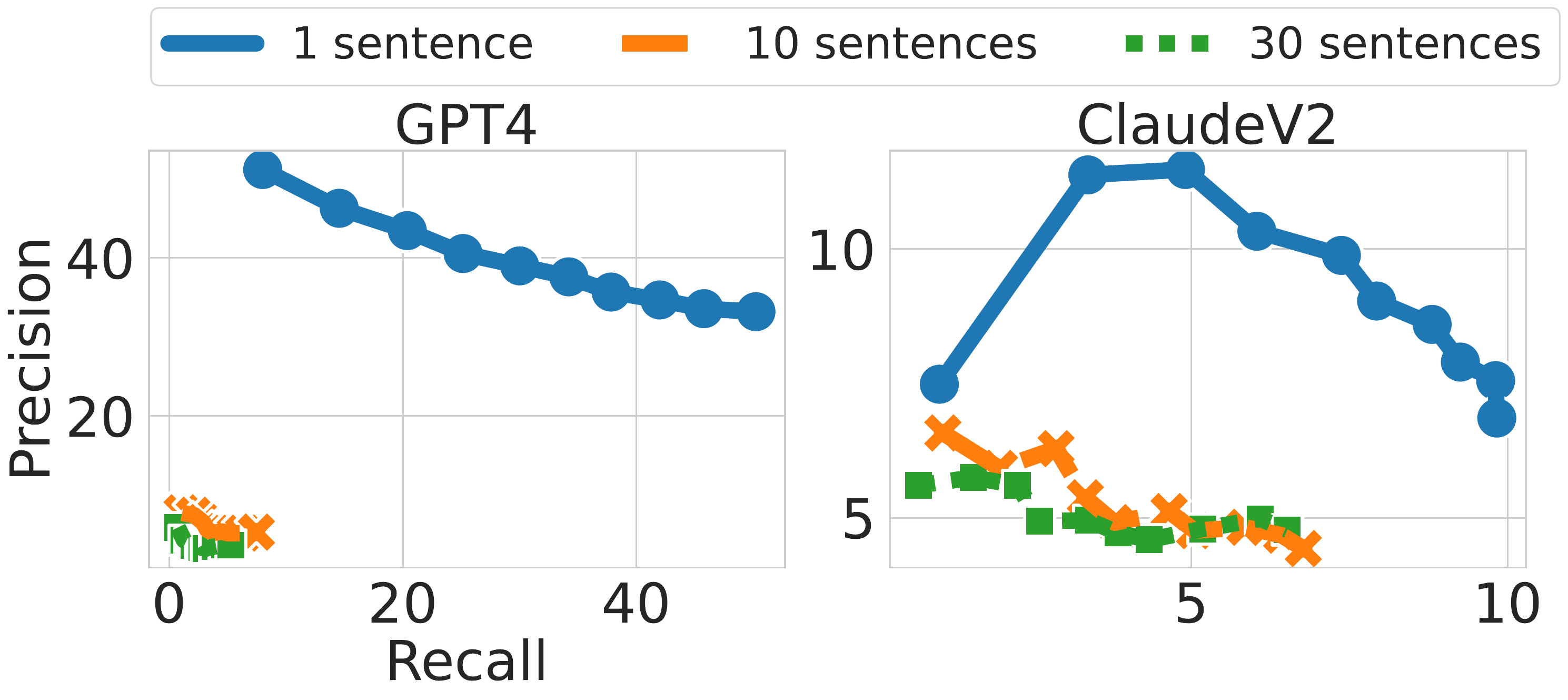}
    \caption{The figures indicate that \GPTFour{} and \ClaudeVTWo{} excel in the task of retrieving 1 sentence, but their performance decreases with the retrieval tasks of 10 sentences and 30 sentences.
    }
    \label{fig:precision-recall-length}
\end{figure}

\paragraph{Calculation of `Random' and `Oracle' Baselines}

In the context of the table above, precision and recall calculations involve two lists of integers: "result" and "golden." In typical precision and recall computations, the "result" list is derived from the models' generations. However, for random calculations, the "result" list consists of integers from 1 to 10. This choice is influenced by our prompt: "NOTE: Only generate an index number without any additional text. For example: 1, 2, 3, 4, 5, 6, 7, 8, 9, 10". Specifically, for challenging tasks, GPT-4 and Claude tend to generate a list ranging from 1 to 10 based on this prompt as default. The random calculation is then performed using this list. In the case of the Oracle calculation, we designate the "result" list to be the same as the "golden" list.

\clearpage

\section{Experiment: Evaluation on Different Stories Lengths for a Fixed Total Context Window}
In our earlier experiments in \S\ref{subsec:scaling-results} and \S\ref{subsec:analogy-results} upon changing the length of each story, we also change the length of the total prompt (i.e., the concatenation of all the stories in the story bank). This essentially creates a confounding two variables that impact the difficulty of the tasks for LMs: (i) length of each story; (ii) the total length of the context. To address this confounding variable, here we fix (ii) and vary (i).

We fix a total context window length budget. Specifically, we fix this budget to be 2K and 1.5K tokens. Then, we fit as many stories that would fit within this total context window budget. The number of the stories that fit in the context window are shown in \autoref{tab:merged_context_story_length_performance}.

\begin{table*}[ht]
    \centering
    \small
    \begin{tabular}{cccccccccc} 
        \toprule
       \multirow{2}{*}{Total Context Length} & \multicolumn{3}{c}{Number of stories} & \multicolumn{3}{c}{\textbf{Scaled} Accuracy} & \multicolumn{3}{c}{Accuracy} \\
       \cmidrule(lr){2-4} \cmidrule(lr){5-7}
       & 1-sent & 10-sent & 30-sent & 1-sent & 10-sent & 30-sent & 1-sent & 10-sent & 30-sent \\
       \midrule
       1500 & 72 & 6 & 3  & 0.04 & 0.15 & 0.07 & 0.05 & 0.29 & 0.38 \\
       2000 & 100 & 10 & 4 & 0.01 & 0.03 & 0.08 & 0.02 & 0.13 & 0.31 \\
       \bottomrule
    \end{tabular}
    \caption{Merged performance metrics for predictions across varying context lengths and story lengths for \TuluTwo{} 70B, with and without scaled accuracy.}
    \label{tab:merged_context_story_length_performance}
\end{table*}

We report the accuracy values for these evaluations, but these values are not comparable to across different length since they have different lower-bounds. For example, a story bank of size 3 leads to a lowerbound of 1/3, while the lowerbound for a story bank with 72 stories is 1/72.

Besides the accuracy metric, we also report a scaled accuracy. 
The scaling is necessary here to make sure that the numbers are all ranged from 0 to 100. To scale a given accuracy value $x$, we can plug it in the following formula: 
$$
\text{scaled-acc} = \frac{x - \text{random-acc}}{1 - \text{random-acc}},
$$
where $\text{random-acc} = 1/(\text{size of story bank})$. 
Overall the results of scaled accuracy values in \autoref{tab:merged_context_story_length_performance} are small. 
Essentially all of our stories of varying length remain difficult, even after accounting for a fixed context window size. 
Whether story length is a stronger factor or the context window length remains somewhat inconclusive and requires more future work.

\newpage

\section{Using Claude-v2 for Story Elaboration}
\label{app:claude-elaboration}

Similar to \S\ref{sec:elaboration}, in \S\ref{subsec:self-generated} we expand a single sentence to craft a story spanning 10 or 30 sentences with \ClaudeVTWo{} this time. 
This directive is similar to how we prompt  \GPTFour{} (example shown in \S\ref{appendix:expansion}), albeit with a slight modification in the guidance given to \ClaudeVTWo{}. 
Below is the instruction:
\vspace{0.4cm}

\begin{tcolorbox}[colback=gray!5!white,colframe=gray!75!black,title=
\textbf{Input:}]
\tt

\textbackslash n\textbackslash n Human:

Given the following sentence, expand it into a 30-sentence story.

Sentence: All that glitters is not gold.

\textbackslash n\textbackslash n Assistant:

\end{tcolorbox}

\begin{tcolorbox}[colback=gray!5!white,colframe=gray!75!black,title=\textbf{Output:}]
\tt 
Here is a 10-sentence story expanding on the sentence "All that glitters is not gold":

Lisa was browsing the jewelry store, looking for a new necklace. So many of the pieces caught her eye with their glittering gems and shiny metals. She tried on a beautiful gold necklace set with bright green emeralds that seemed to glow. Although it was very pretty, when she looked at the price tag she gasped - it cost a small fortune! Lisa realized that just because something glitters, doesn't mean it's as valuable as it appears. She put the necklace back and continued browsing for something more affordable. In a corner case, she found a simple silver chain with a small pearl pendant. It didn't sparkle as brightly, but she knew pearls were classic and elegant. Lisa tried on the necklace and loved how it looked on her. She decided it was perfect, and happily purchased it for a fraction of the price of the emerald necklace. Though it didn't glitter quite as much, she knew this beautiful pearl pendant was the true treasure.

\end{tcolorbox}

\end{document}